\documentclass[11pt,letterpaper]{article}
\usepackage[letterpaper]{geometry}
\usepackage{acl2012}
\usepackage{times}
\usepackage{latexsym}
\usepackage{amsmath}
\usepackage{amssymb}
\usepackage{multirow}
\usepackage{url}
\usepackage{color,soul}
\usepackage{lipsum}
\usepackage{alltt}
\usepackage{pifont}
\usepackage{booktabs}

\usepackage{graphicx}
\usepackage[export]{adjustbox}
\usepackage{multirow}
\usepackage{subcaption}
\title{Separating Answers from Queries for Neural Reading Comprehension}

\author{Dirk Weissenborn \\
	    Language Technology Lab, DFKI \\
	    Alt-Moabit 91c \\
	    Berlin, Germany \\
        {\tt dirk.weissenborn@dfki.de}}

\date{}

\begin{document}
\maketitle
\begin{abstract}
We present a novel neural architecture for answering queries, designed to optimally leverage explicit support in the form of query-answer memories. Our model is able to refine and update a given query while separately accumulating evidence for predicting the answer. Its architecture reflects this separation with dedicated embedding matrices and loosely connected information pathways (modules) for updating the query and accumulating evidence. This separation of responsibilities effectively decouples the search for query related support and the prediction of the answer. On recent benchmark datasets for reading comprehension, our model achieves state-of-the-art results. A qualitative analysis reveals that the model effectively accumulates weighted evidence from the query and over multiple support retrieval cycles which results in a robust answer prediction.
\end{abstract}

\section{Introduction}

Recent advances in many NLP tasks were achieved by utilizing neural architectures that employ some form of external memory. Making use of explicit memories enables these models to bridge long-range dependencies and solve more complex reasoning tasks that might involve multiple observations. Neural architectures equipped with explicit memories are able to achieve impressive results on a variety of NLP tasks. Memory Networks \cite{weston2014memory,sukhbaatar2015end}, for example, are able to answer questions which require a higher level of reading comprehension and possibly reason over multiple observations. 

The use of some form of external memory appears essential when tackling complex queries that require comprehension of a given context (support). The memory module stores explicit, contextual information of the support which either contains the correct answer or clues that can lead to it. For instance, attention-based architectures \cite{hermann2015teaching} encode supporting contexts typically with (bi-directional) recurrent neural networks (RNN) into $h$-dimensional latent representations (\textit{hidden states}), which jointly serve as a form of memory. End-to-end Memory Networks \cite{sukhbaatar2015end} are similar although they split the support into individual parts that are separately encoded to form \textit{memories}. These systems utilize the same learned query representation both for selecting memories (matching) and predicting the actual answer (prediction) which can affect the overall performance. Recent work successfully addressed this issue by directly using the retrieved hidden states \cite{cheng2016long} or the attention weights \cite{kadlec2016text} as pointers to the answer \cite{vinyals2015pointer} for prediction.

In this work we propose a novel end-to-end neural architecture for answering queries. It explicitly separates queries from answers which is reflected in the representation of supporting knowledge as query-answer pairs and in the general architecture. In particular, we employ dedicated embedding matrices and loosely connected information pathways (modules) for updating the query and answer representation. This separation of responsibilities increases the capabilities of the model to search through the support while selectively accumulating evidence for the answer in parallel. The representation of the support reflects the task of answering queries directly which facilitates its utilization by the model. We evaluate our approach on two reading comprehension tasks that involve answering cloze-style \cite{taylor1953cloze} queries, namely the CNN/DailyMail QA task \cite{hermann2015teaching} and the named entity (NE) subtask of the Children's Book Test (CBT) \cite{Hill2015TheGP}. These datasets provide only one document as support per query but this is not a restriction because our model can also handle multiple documents. Our contributions are the following: i) we introduce a new representation of supporting memories in form of query-answer pairs (\S\ref{sec:support_kn}) based on which ii) we develop a neural architecture for answering queries that leverages this representation (\S\ref{sec:qa_network}), iii) we evaluate our system on two reading comprehension benchmark datasets against other competitive systems achieving state-of-the-art results (\S\ref{sec:results}), and iv) we give insights into the systems ability to utilize multiple support retrieval cycles for improving its reading comprehension performance (\S\ref{sec:results_varying_hops} and \S\ref{sec:qualitative}).

\section{Related Work}

Utilizing explicit memory in end-to-end differentiable neural architectures has enabled models to solve complex tasks that require learning simple algorithms, or processing and reasoning over large amounts of contextual information. Traditional architectures, such as RNNs like the LSTM \cite{hochreiter1997long} or GRU \cite{chung2014empirical}, are not suited for these kind of tasks due to their limited memory capacity and difficulties to learn long-range dependencies in large contexts.

\newcite{graves2014neural} introduced Neural Turing Machines (NTM). NTMs augment traditional RNNs with external memory that can be written to and read from. The memory is composed of a predefined number of writable slots. They are addressable via content or position shifts which allows solving simple algorithmic tasks. The capacity is also limited, but the external memory slots can carry information over long ranges more easily than traditional RNNs. NTMs inspired subsequent work on using different kinds of external memory, like queues and stacks for solving transduction tasks \cite{grefenstette2015learning} or neural theorem provers to perform first-order \cite{rocktaschel2016learning} inference.

Attention-based architectures store information, typically in form of hidden RNN states, dynamically for each time-step while processing a given context. These states are retrieved through an attention mechanism that softly selects a state that matches a given query state. This can be viewed as keeping the encoded context in memory. Such architectures achieved impressive results on tasks that involve larger contexts such as Reading Comprehension \cite{hermann2015teaching,kadlec2016text,chen2016thorough}, Machine Translation \cite{bahdanau2014neural,luong2015effective} or recognizing textual entailment \cite{rocktaschel2015reasoning,cheng2016long}.

Based on ideas of the attention mechanism, End-to-end Memory Networks \cite{sukhbaatar2015end} select explicit memories for query answering. Memories are encoded into two representations: i) the input representation for query matching and ii) the output representation for subsequent utilization. This distinction is important, because the representation that is used to match the original query has a different responsibility than the representation that is used to answer or update the query. Thus, attention-based approaches for answering queries using supporting documents can be considered a special case of Memory Networks where hidden states form both the input- and output representation of the individual memories which are jointly encoded. Variants of Memory Networks have achieved very good results in various NLP tasks, such as language modeling \cite{sukhbaatar2015end}, reading comprehension \cite{Hill2015TheGP} and question answering \cite{sukhbaatar2015end,Kumar2015AskMA,miller2016key}. 

One important contribution of Memory Networks is the idea of refining or updating the query \cite{sukhbaatar2015end} or memories \cite{Kumar2015AskMA} for multiple memory retrieval cycles before answering the query. This idea lead to significant improvements for architectures that employ the attention mechanism iteratively for reading comprehension tasks \cite{sordoni2016iterative}.

Recently, other forms of explicit memory have been suggested for neural architectures. For instance, associative memory can be used to effectively compress multiple memories into redundant copies of a single memory array. It has shown very promising results, e.g., for language modeling \cite{danihelka2016associative} or recognizing textual entailment \cite{weissenborn2016dual}, and might therefore be suited to compress large amounts of external memories when used in conjunction with our model.

\section{Query-Answer Neural Network}\label{sec:qa_network}

\begin{figure*}[t]
    \centering
    \includegraphics[width=\textwidth]{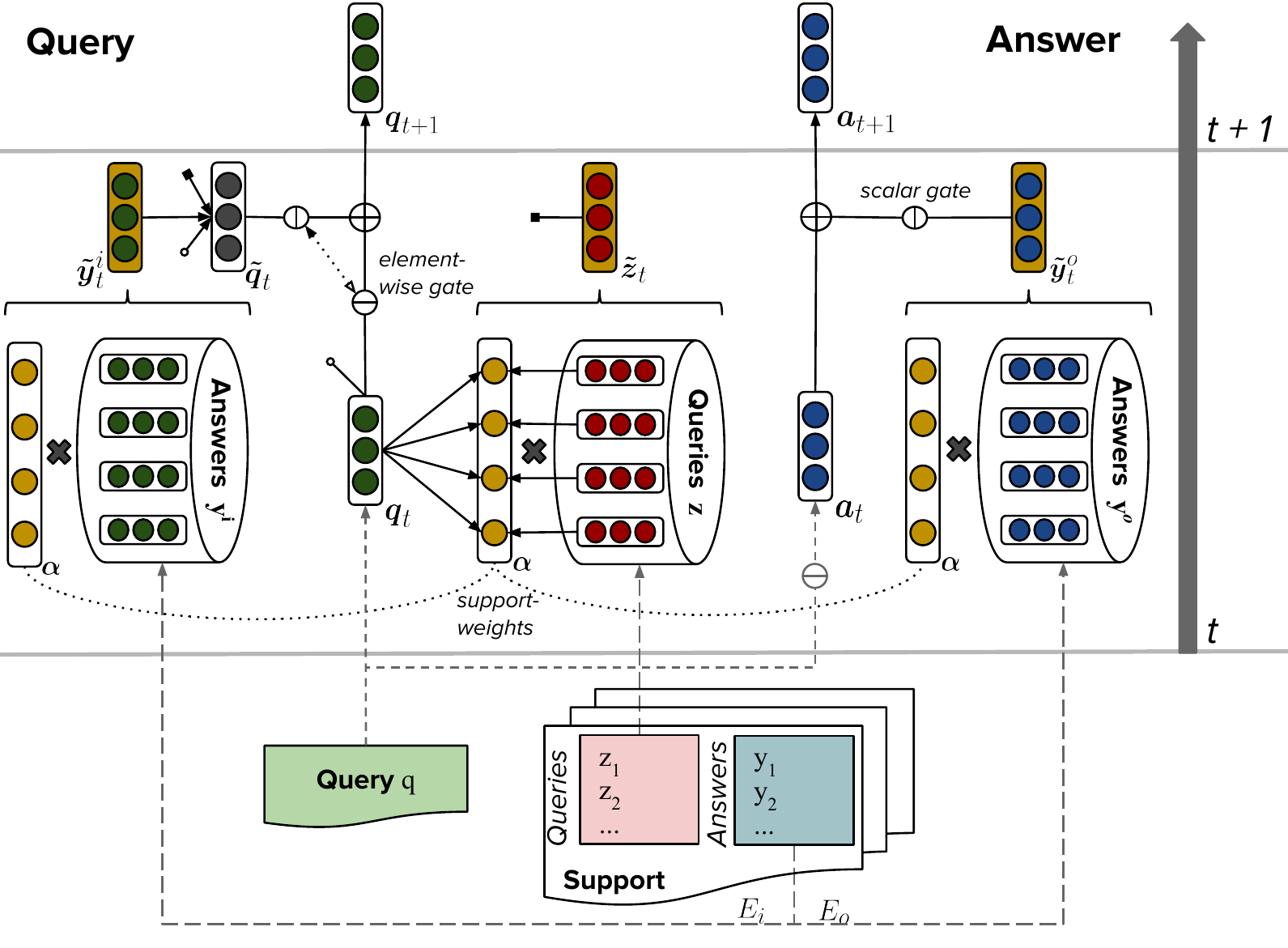}
    \caption{Illustration of our architecture which demonstrates an support retrieval cycle (hop) along with its corresponding update of the query and answer utilizing supporting queries and answers ($(z,y)$-pairs). The query representation is initialized ($t=0$) by encoding the query string $q$. The initial answer representation is computed based on the initial query representation.}
    \label{fig:qa_network}
\end{figure*}
Our query-answer neural network utilizes supporting knowledge in the form of explicit query-answer pairs $(z,y)$ to predict the answer $a$ to a given query $q$. Answers  from support ($y$) are weighted via matching scores between their corresponding support query $z$ and the actual query $q$. Once a weighted query-answer pair $(\tilde{z},\tilde{y})$ has been retrieved, it is used to update the current query and the predicted answer representation for a subsequent support retrieval cycle (hop). This process can be repeated for a specified number of hops ($T$). Finally, we use the predicted answer representation after $T$ hops as input for answer classification given a set of possible answer candidates. Note that this approach does not require supporting answers to correspond to the answer candidates.

\subsection{Supporting Knowledge}\label{sec:support_kn}
Our model stores supporting knowledge as pairs of queries ($z$) and answers ($y$). Given a set of supporting documents, $(z,y)$-pairs are formed by i) detecting task-specific \textit{spans-of-interest} (SOIs), ii) forming $(z,y)$-pairs for each SOI. In this work we consider cloze-style $(z,y)$-pairs. Thus, given a $(z,y)$-pair $z$ corresponds to an entire document with a gap at a particular SOI (cloze-text) and $y$ to the filler of this gap. Consider the following example:
\begin{alltt}
\textbf{Query: }
q: Schweinsteiger plays for the 
   national team of \underline{\quad}   
\textbf{Support Document 1: }
D: Schweinsteiger scored against 
   \underline{Ukraine}
\textbf{Support Document 2: }
D: \underline{Germany} played against \underline{Ukraine}
\end{alltt}

From this example we extract the following supporting query-answer pairs, if we identified all countries (underlined) as SOIs:

\begin{alltt}
\textbf{Support QA 1:}
z: Schweinsteiger scored against \underline{\quad}
y: Ukraine
\textbf{Support QA 2:}
z: \underline{\quad} played against Ukraine
y: Germany
\textbf{Support QA 3:}
z: Germany played against \underline{\quad}
y: Ukraine
\end{alltt}

Note that spans-of-interest can cover almost anything, e.g., entire sentences or only single words. Defining SOIs and their respective answers can be adapted to the needs of the task at hand.

\subsection{Encoding Queries}\label{sec:query_encoding}

Given a (supporting) document $D=(x_1, ..., x_N)$ of symbols and spans-of-interest (SOIs) $P=\{(l_s,l_e) \,\, | \,\, l_s, l_e \in [1, N]\}$, at first all symbols $x_l$ are embedded by an embedding matrix $E_i$. Next, the entire document $D$ is encoded by a bi-directional RNN resulting in representations $\boldsymbol{h}^f_l \in \mathbb{R}^{h}$ of the forward-RNN and $\boldsymbol{h}^b_l \in \mathbb{R}^{h}$ of the backward-RNN for each document position $l \in [1, N]$. Afterwards, we form the following query representation for each $(l_s,l_e) \in P$:

\begin{align}
\boldsymbol{z}^l =& W_q \begin{bmatrix}\boldsymbol{h}^f_{l_s-1} \\ \boldsymbol{h}^b_{l_e+1} \end{bmatrix}  \quad W_q \in \mathbb{R}^{h \times 2h}
\end{align}

The trainable parameter-matrix $W_q$ is initialized with $[I^n; I^n]$ and additional random noise, where $I^n$ is the identity matrix. Thus, initially $\boldsymbol{z}^l$ corresponds roughly to the sum of the forward state $\boldsymbol{h}^f_{l_s-1}$ of the left context and the backward state $\boldsymbol{h}^b_{l_e+1}$ of the right context. In order to ensure that the query representation only considers the outer context of the respective SOI it is required that $l_s \le l_e$.\footnote{It is also possible to define $l_s > l_e$, s.t. the span-of-interest becomes part of the query which might be important for some tasks. However, we are not considering these types of tasks in this work.} Encoding supporting queries in this way has the advantage, that the entire context is encoded in contrast to restricting the context to a fixed-size context window or sentence. Furthermore, word-order and positional information is captured naturally by employing RNNs. Encoding the actual query $q$ is identical to the encoding supporting queries $z$. 

\subsection{Encoding Answers}\label{sec:answer_encoding}
In this work, we consider answers to be individual symbols. However, this approach can be extended to sequences of symbols as well.

\paragraph{Answer Candidates} Answer candidates $c \in A_q$ for a query $q$ are embedded ($\boldsymbol{c}$) by a second embedding matrix $E_o$. 

\paragraph{Supporting Answers} For cloze-style queries supporting answers ($y$) correspond to the symbols at SOIs within the support.\footnote{Note, supporting answers do not necessarily have to correspond to an actual answer candidate.} There are two encodings of $y$ with different applications: i) its corresponding output embedding $\boldsymbol{y}^o$ from $E_o$, and ii) its corresponding input embedding $\boldsymbol{y}^i$  from $E_i$. $\boldsymbol{y}^o$ is used to update the current answer of the model and $\boldsymbol{y}^i$ is used to update the query. 

The intuition behind using $\boldsymbol{y}^i$ for updating the query representation is that we want to use the answer as if it was a word of the original query and would thus be embedded by $E_i$. $\boldsymbol{y}^o$ corresponds to the embeddings of $E_o$ that are only used for answer prediction.

It is possible to fix $E_o$ to the identity matrix which results in the accumulation of support weights as score for each answer candidate. This is similar to using the attention weights for answer scoring as described in \newcite{kadlec2016text}.

\subsection{Supporting Answer Retrieval}
A supporting answer $\tilde{y}$ is selected softly from all ($M$) supporting $(z,y)$-pairs by $\operatorname{softmax}$-weights based on similarity scores between all $\boldsymbol{z}_k$ and the query $\boldsymbol{q}$. These support weights can be viewed as attention weights over the respective supporting $(z,y)$-pairs.

\begin{align}
\alpha_k &= \frac{\operatorname{exp}(\boldsymbol{q} \cdot \boldsymbol{z}_k)}{\sum_{k^\prime=1}^M\operatorname{exp}(\boldsymbol{q} \cdot \boldsymbol{z}_{k^\prime})}  \\
\begin{bmatrix} \boldsymbol{\tilde{y}}^i \\ \boldsymbol{\tilde{y}}^o \\ \boldsymbol{\tilde{z}} \end{bmatrix} &= \sum_{k=1}^M \alpha_k \begin{bmatrix} \boldsymbol{y}^i_k \\ \boldsymbol{y}^o_k \\ \boldsymbol{z}_k \end{bmatrix} \label{eq:zy}
\end{align}

\subsection{Query \& Answer Update}\label{sec:multi_hop}
The query representation $\boldsymbol{q}$ and predicted answer representation $\boldsymbol{a}$ are consecutively updated by using supporting $(z,y)$-pairs realizing multi-hop support retrieval. For instance, in the example of \S\ref{sec:support_kn} the model might find \textit{Support QA 1} to fit the original query best and retrieve the (wrong) answer (\textit{Ukraine}). It is reasonable to update the original query with \textit{Support QA 1} which includes the answer \textit{Ukraine}. The subsequent, updated query eventually leads to the correct answer \textit{Germany} of \textit{Support QA 2}. Figure~\ref{fig:qa_network} illustrates this process.

We utilize the weighted support-queries $\boldsymbol{\tilde{z}}_t$ and their corresponding answer input-representations $\boldsymbol{y}^i_t$ (Eq.~\ref{eq:zy}) to update the current query $\boldsymbol{q}_t$ by an element-wise weighted addition (Eq.~\ref{eq:query_update}), where $\boldsymbol{q}_0 = \boldsymbol{q}$ (the encoded query).

\begin{align}
\boldsymbol{\tilde{q}}_t &= \tanh \left ( U^q_c \begin{bmatrix}\boldsymbol{q}_t \nonumber\\ \boldsymbol{\tilde{y}}^i_t \\ \boldsymbol{\tilde{z}}_t \end{bmatrix} \right ) \\
\boldsymbol{g}^q_t &= \operatorname{sigmoid} \left ( U^q_g \begin{bmatrix}\boldsymbol{q}_t \\  \boldsymbol{\tilde{z}}_t \end{bmatrix} + \boldsymbol{b}^q_g \right ) \nonumber \\
\boldsymbol{q}_{t+1} &= \boldsymbol{g}^q_t \odot \boldsymbol{q}_t + (1-\boldsymbol{g}^q_t) \odot \boldsymbol{\tilde{q}}_t \label{eq:query_update}
\end{align}

The answer is initialized by a gated linear transformation of the initial query $\boldsymbol{q}_0$ (Eq.~\ref{eq:answer_init}). The query-answer-gate $g^a_q$ decides whether the query itself can be utilized to infer the answer for a specific task or not. The answer representation at hop $t$ represented by $\boldsymbol{a}_t$ is updated to $\boldsymbol{a}_{t+1}$ by adding the gated, retrieved answer $\boldsymbol{\tilde{y}}^o_t$ (Eq.~\ref{eq:answer_update}). The scalar answer accumulation gate $g^a_t$ (Eq.~\ref{eq:answer_gate}) depends on: i) the similarity between the current query $\boldsymbol{q}_t$ and the weighted support queries $\boldsymbol{\tilde{z}}_t$, ii) the similarity of the original query encoded as answer $\boldsymbol{a}_0$ and the weighted support answer representation $\boldsymbol{\tilde{y}}^o_t$ retrieved from support and iii) $\eta_t$ which measures the highest answer candidate probability if $\boldsymbol{\tilde{y}}^o_t$ was the final answer representation (Eq.~\ref{eq:discriminative} which refers to \S\ref{sec:scoring}).

\begin{align}
\boldsymbol{a}_0 &= \operatorname{sigmoid} (g^a_q) \, U^a_q \boldsymbol{q} \label{eq:answer_init} \\
\boldsymbol{a}_{t+1} &= \boldsymbol{a}_t + g^a_t \boldsymbol{\tilde{y}}^o_t \label{eq:answer_update} \\
g^a_t &= \operatorname{sigmoid} \left ( \boldsymbol{u}^a_g \begin{bmatrix}  \boldsymbol{q}_t \odot \boldsymbol{\tilde{z}}_t\\ \boldsymbol{a}_0 \odot \boldsymbol{\tilde{y}}^o_t \\ \eta_t 
\end{bmatrix} + b_a \right ) \label{eq:answer_gate}  \\
\eta_t &= \underset{c \in A_q}{\max}\; p_{q}(\boldsymbol{c} | \boldsymbol{\tilde{y}}^o_t) \label{eq:discriminative} 
\end{align}

The trainable parameters of this module have the following dimensions: $U^q_c \in \mathbb{R}^{h\times3h}; \, U^q_g \in \mathbb{R}^{h \times 2h};\, U_a \in \mathbb{R}^{h \times h};\,\boldsymbol{b}^{\cdot}_g, \boldsymbol{u}^a_g \in \mathbb{R}^{h}; \, g^a_q, u_a, b_a \in \mathbb{R}$.

\subsection{Answer Scoring \& Training}\label{sec:scoring}
After a maximum number of hops $T$, scores $s_{q,c}$ of all answer candidates $c \in A_q$ are calculated using the inner product between their respective embeddings $\boldsymbol{c}$ and the final answer representation $\boldsymbol{a}_T$:

\begin{align}
\forall c \in A_q: s_{q}(\boldsymbol{a}, \boldsymbol{c}) &= \boldsymbol{a} \cdot \boldsymbol{c} \nonumber \\
p_{q}(\boldsymbol{c}| \boldsymbol{a}) &= \frac{e^{s_{q}(\boldsymbol{a}, \boldsymbol{c})}}{\sum_{c^\prime \in A_q} e^{s_{q}(\boldsymbol{a}, \boldsymbol{c}^\prime)}} \label{eq:final_prob}
\end{align}

Finally, the model is trained by minimizing the cross-entropy loss using the $\operatorname{softmax}$-weights (Eq.~\ref{eq:final_prob}) of candidate scores as the predicted probabilities.

\section{Experiments}

\subsection{Setup}

\paragraph{Dataset}
We evaluate our architecture on two recently proposed benchmarks for reading comprehension. Both benchmarks require a system to answer a cloze-style query solely based on a single supporting document. \newcite{hermann2015teaching} created two datasets (CNN, DailyMail) from news articles. For each article, queries were created from their respective summaries by removing a named entity from the summary sentence that has to be predicted. All articles in the dataset are pre-processed by named entity recognition, co-reference resolution and entity anonymization. Similar in mind, \newcite{Hill2015TheGP} created the \textit{Children's Book Test} (CBT). For this dataset, passages of children's books of $21$ sentences were extracted. Within the last sentence of the passage a word is removed that has to be predicted. The dataset is split into subtasks depending on the part-of-speech tag of the word that has to be predicted. We evaluate our model on the named entity (NE) subtask because it is the most challenging subtask for traditional language models.

\paragraph{Input Presentation \& Encoding} 
The input to the model consists of the context document and the query. The actual query is the cloze-text for the position of the removed named entity, which is replaced by a placeholder symbol. The entire input (document + query) is encoded by a bi-directional GRU from which the query and answer representations are computed as described in \S\ref{sec:query_encoding} and \S\ref{sec:answer_encoding}. Supporting $(z,y)$-pairs are extracted at occurrences of an answer candidate (all entities for CNN/DailyMail, given for CBT) in the context document in form of cloze-text (query) and its corresponding filler (answer).

\paragraph{Training}
For all experiments, we use a hidden dimension (and embedding-size) of $h=256$. We train models with and without pre-trained word vectors. The input embedding matrix $E_i$ is partially initialized with $100$-dimensional \verb|Glove|-embeddings \cite{pennington2014glove} and randomly for the rest ($156$ dimensions) when using pre-trained word vectors. In general, embeddings are initialized with a Gaussian of  $0$-mean and $0.1$-stddev, matrices as described in \newcite{glorot2010understanding} and biases with $0$, except for the encoder GRU update-gate bias which is initialized with $1$. Dropout is applied with a rate of $0.2$ to the embedded input words for regularization. We train our system using mini-batch SGD with ADAM \cite{kingma2014adam} for optimization using an initial learning-rate of $0.001$ that is halved whenever the accuracy on the development set drops between checkpoints and the the first entire epoch has passed. If the accuracy drops between entire epochs training is stopped. The mini-batch sizes/respective checkpoint iterations are $128$/$500$ for the DailyMail and CNN datasets\footnote{batch-size is reduced to $64$ for ensembles due to memory constraints}, and $32$/$1000$ for the CBT NE dataset. We trained both single models and ensembles of 5 models. Note, that similar to \newcite{chen2016thorough}, we do not consider all words but only entities as answer candidates for the CNN/DailyMail dataset. Our models are trained with $4$ and $8$ consecutive support retrieval cycles (hops) from the support, which performed better than using only $1$ or $2$. All models were implemented with TensorFlow \cite{tensorflow2015}.

\subsection{Empirical Results} \label{sec:results}
\begin{table*}[t]
\footnotesize
\begin{subtable}{.6\textwidth}
\centering
\begin{tabular}{l c c c}
\toprule
\textbf{Model} & \textbf{CBT NE} & \textbf{CNN} & \textbf{DailyMail} \\
\midrule
Attentive Reader$^1$ & - & $63.0$ & $69.0$ \\
Stanford Attentive Reader (Glove)$^2$ & - & $\underline{72.4}$ & $\underline{75.8}$ \\
\midrule
EpiReader$^{3}$ (answer re-ranking)& $\underline{69.7}$ & $\underline{\boldsymbol{74.0}}$ & -  \\
\midrule
\multicolumn{4}{l}{\textit{Multi-hop Systems}} \\
\midrule
Impatient Reader$^1$ & - & $63.8$ & $68.0$\\
MemNNs (window)$^4$ & $49.3$ & $60.6$ & - \\
MemNNs (window + self-sup.)$^4$ & $66.6$ & $66.8$ & - \\
GA Reader$^{5}$ & $\underline{69.0}$ & $\underline{73.8}$ & $\underline{75.7}$ \\ 
Iterative Attentive Reader$^{6}$ & $68.6$ & $73.3$ & - \\
\midrule
QANN (4 Hops) & $69.4$ & $72.6$ & $\underline{76.6}$ \\
QANN (8 Hops) & $\underline{70.3}$ & $\underline{73.4}$ & $76.4$ \\
\midrule
QANN (4 Hops, Glove) & $\underline{\boldsymbol{70.6}}$ & $\underline{73.7}$ & $76.9$ \\
QANN (8 Hops, Glove) & $\underline{\boldsymbol{70.6}}$ & $73.6$ & $\underline{\boldsymbol{77.2}}$\\
\midrule
\midrule
\multicolumn{4}{l}{\textit{Ensembles}} \\
\midrule
EpiReader$^{3}$ (answer re-ranking)& $71.8$ & - & -  \\
MemNNs (window + self-sup.)$^4$ & $66.6$ & $68.4$ & - \\
GA Reader$^{5}$ & $71.9$ & $\boldsymbol{\underline{77.4}}$ & $\underline{78.1}$ \\ 
Iterative Attentive Reader$^{6}$ & $\underline{72.0}$ & $76.1$ & - \\
\midrule
QANN (4 Hops, Glove) & $\boldsymbol{\underline{72.9}}$ & $\underline{76.8}$ & $\boldsymbol{\underline{79.2}}$ \\
\bottomrule
\end{tabular}
\caption{}
\label{tab:results}
\end{subtable}~
\begin{subtable}{.4\textwidth}
    \centering
    \begin{tabular}{c c c}
        \toprule
        \textbf{Hops} & \textbf{CBT NE} & \textbf{CNN} \\
        \midrule
        $1$ & $60.4$ &$67.0$ \\
        $2$ & $67.0$ &$72.0$ \\ 
        $3$ & $70.0$ &$73.3$ \\
        \midrule
        $4$ & $70.6$ & $73.7$ \\
        \midrule
        $5$ & $\boldsymbol{71.2}$ & $73.3$ \\
        $6$ & $71.0$ & $\boldsymbol{73.8}$ \\ 
        $7$ & $70.7$ & $73.2$ \\
        \bottomrule
    \end{tabular}
    \caption{}
    \label{tab:results_hops}
\end{subtable}
\caption{(a) Accuracies of different models on 3 benchmark test datasets for reading comprehension. \newcite{hermann2015teaching}~$^1$, \newcite{chen2016thorough}~$^2$, \newcite{trischler2016natural}~$^3$, \newcite{Hill2015TheGP}~$^4$, \newcite{dhingra2016gated}~$^5$, \newcite{sordoni2016iterative}~$^6$. Note, that the works $^{3, 5, 6}$ are very recent. (b) Accuracies of QANNs on the CBT NE and CNN testsets when employing a model trained with 4 support retrieval cycles, but applied on a varying number of retrieval cycles (hops). }
\end{table*}

The results of our model (QANN) on the two benchmarks are presented in Table~\ref{tab:results}. They show that our trained single models and ensembles achieve state-of-the-art results on the CBT NE and the DailyMail dataset and perform similar to recent work on the CNN dataset. An important observation is that our model outperforms the Memory Networks by a large margin. Even with self-supervision, which explicitly introduces a training objective for selecting the correct memory at each hop, Memory Networks are clearly outperformed by our system. We attribute this to the query-answer representation of our supporting memory and the related architectural changes that separate end-to-end Memory Networks from QANNs.

For our models we observe that using $4$ instead of $8$ support retrieval cycles (hops) makes a difference only if \verb|Glove| is not used for initialization. Using \verb|Glove| to (partially) initialize embeddings gives a boost in performance on all datasets. 

We would like to point out that the only systems with comparable results to ours use either the attention-weights over context-words (Iterative- and Gated Attentive Reader, EpiReader, inspired by \newcite{kadlec2016text}) or the retrieved hidden state (Stanford Attentive Reader) to predict the final answer. Those works follow a similar idea as this work which separates the answer that is used for prediction from the query. 

The EpiReader \cite{trischler2016natural} achieves slightly better results on the CNN dataset, because it employs a second neural network (\textit{reasoner}) that re-ranks the output of an attention based model to improve prediction. This idea is orthogonal to our work and can be used to improve QANNs similarly.

\begin{figure*}[p]
    \centering
    \begin{subfigure}[t]{\textwidth}
    \includegraphics[width=\textwidth]{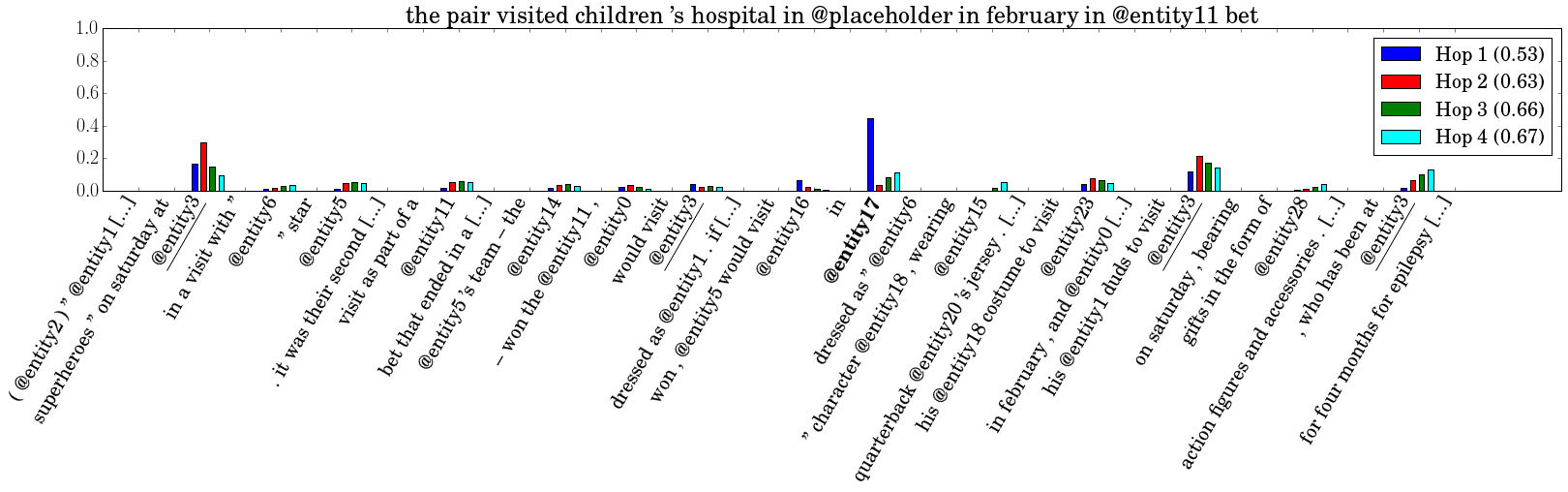}
    \caption{CNN: Correct answer retrieved after first hop but wrong prediction after subsequent hops.}\label{fig:example_1}
    \end{subfigure}

    \vspace{1cm}
    
    \begin{subfigure}[t]{\textwidth}
    \includegraphics[width=\textwidth]{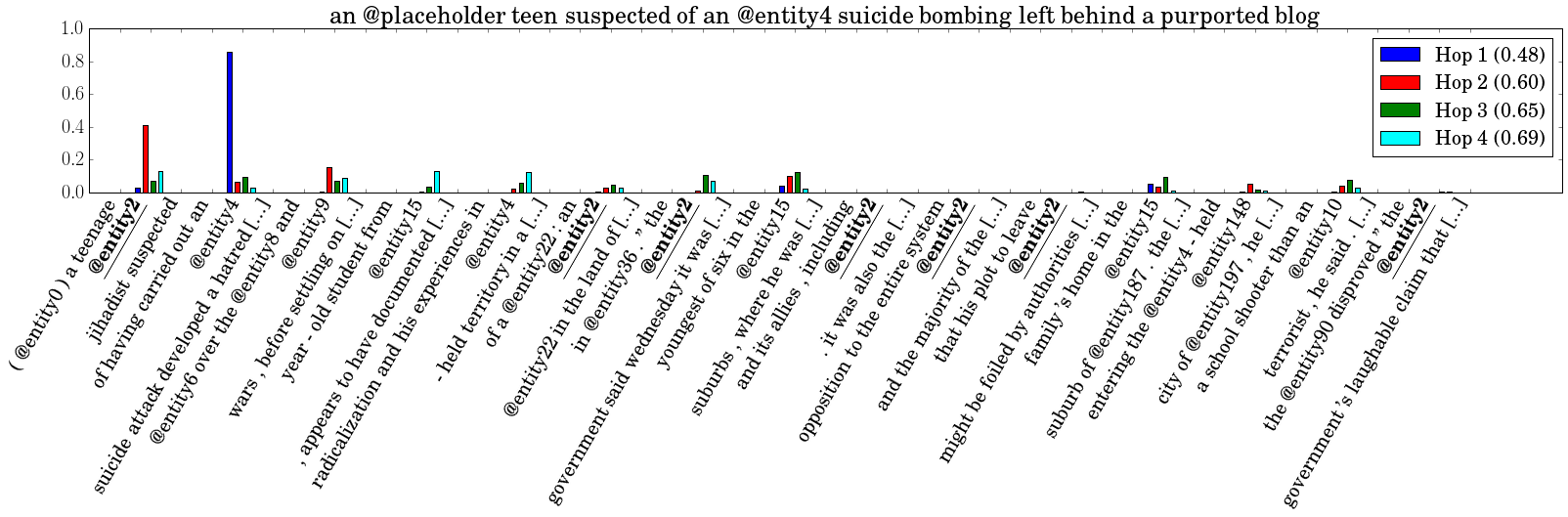}
    \caption{CNN: Wrong answer retrieved after first hop but corrected in subsequent hops.}\label{fig:example_2}
    \end{subfigure}
    
    \vspace{1cm}
    
    \begin{subfigure}[t]{\textwidth}
    \includegraphics[width=\textwidth]{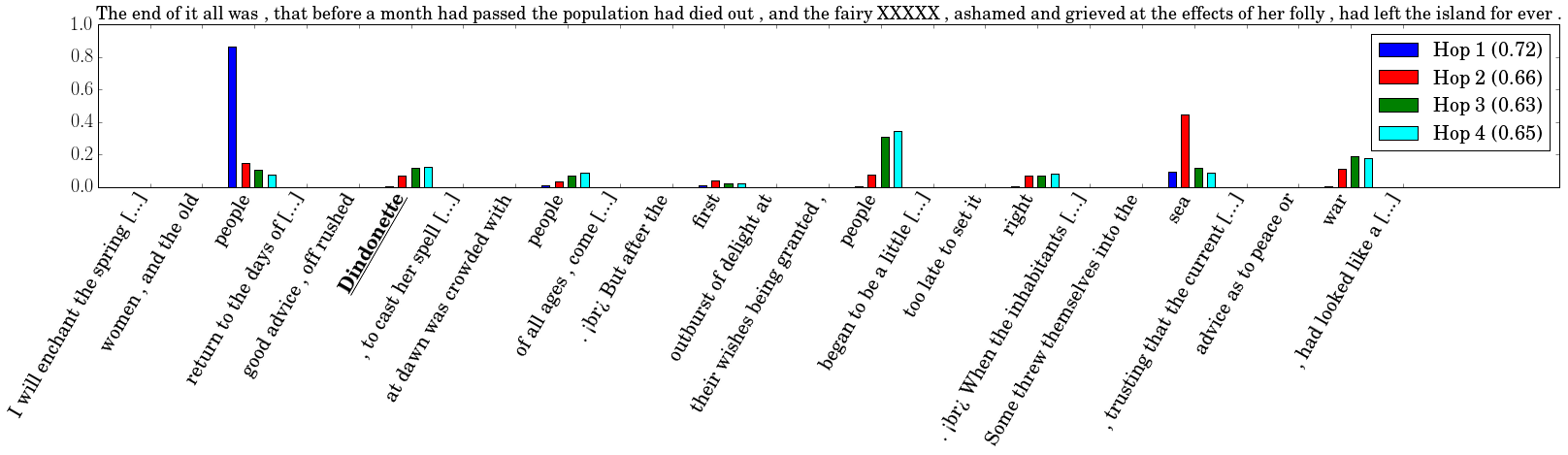}
    \caption{CBT NE: Retrieved answers are wrong yet system predicts answer correctly.}\label{fig:cbt_example}
    \end{subfigure}
    
    \caption{Examples of support (attention) weights for each hop for models trained with fixed 4 hops. The legends show the activity of the respective answer gates for each hop in brackets. The predicted answer is underlined and the correct answer is displayed in bold-face.} \label{fig:examples}
\end{figure*}

The Iterative Attentive Reader \cite{sordoni2016iterative} alternates between attention on specific parts of the query and attention on the context document. The authors found that attending over the query is very useful, however, the attention is usually set on the placeholder symbol which is similar in our approach. They achieve similar results on the CNN dataset but are outperformed by our models on the CBT NE dataset. We attribute this improvement to our answer update mechanism (\S\ref{sec:multi_hop}) which accumulates dedicated answer embeddings from support over multiple hops and from the original query through gates. Thus, the final answer prediction depends not only on the attention (support) weights, but also on the query itself and the answer embeddings. This advantage cannot be exploited as much in the CNN and DailyMail datasets because entities and thus all answers are anonymized. As illustration, Figure~\ref{fig:cbt_example} provides an example of a correctly predicted answer that does not align with the computed support weights (see \S\ref{sec:qualitative} for more details).

\subsection{Impact of Multiple Answer Retrievals}\label{sec:results_varying_hops}

We trained our models with different numbers of support retrieval cycles (hops). We found that using at least $4$ hops leads to a significantly better performance than using only $1$ or $2$ hops. This indicates that multiple consecutive support retrievals and respective query updates are important for a robust performance on the reading comprehension tasks.

In addition, we evaluate the differences in performance when varying the number of hops when using our model trained on $4$ hops. This evaluation gives insights into accuracy gains and the stability of answer prediction when increasing the number of hops. The results presented in Table~\ref{tab:results_hops} demonstrate that the model gains most until $3$ hops, after which results are quite stable. The most pronounced difference occurs between using only $1$ and $2$ hops. The relative stability of performance between $3$ to $7$ hops indicates that the system learns to utilize the gating mechanisms which decide to keep or update the current query and accumulate answers successfully. Even though our model was trained with 4 hops, the best results for the CBT NE and CNN testsets were achieved when utilizing the model with additional hops (5 for the CBT NE and 6 for CNN). Surprisingly, for CBT we found a rather large improvement of about 0.6 percentage points in accuracy.

\subsection{Analysis}\label{sec:qualitative}

In a qualitative analysis of our system on sampled documents from the CNN and CBT NE dataset, we found that the correct answer is retrieved already after the first hop and kept until prediction in most cases (64\% on 1000 sampled CNN examples). However, there are interesting exceptions to this rule that are displayed in Figure~\ref{fig:examples}. It shows example documents with respective attention weights for supporting spans-of-interest (positions of answer candidates) in each hop. A general observation is that the support weights are very pronounced for the first hop and spread over an increasing number of positions with each additional hop. We find that highly weighted positions can vary significantly between hops which shows that the query is updated. As we have shown empirically in \S\ref{sec:results_varying_hops} this can have a positive effect (8.4\%), see for example Figure~\ref{fig:example_2}. However, sometimes this can also result in an incorrect prediction, although the answer was correctly found in the first hop (2.7\%) as demonstrated in Figure~\ref{fig:example_1}.

A very interesting example from the CBT NE validation set is displayed in Figure~\ref{fig:cbt_example}. It shows that the system puts high support weights on different positions in the document, but never on the correct answer. Nevertheless, to our surprise the model predicts the answer correctly anyway. One explanation might be that the model has learned that general words like ``people" or ``sea" are not good answers for the CBT NE dataset (e.g., through answer embeddings with small norm). Another explanation is that the query itself (which is also used to form the final answer representation) puts a strong bias on the final answer. To test the latter hypothesis we set the query-gate of Eq.~\ref{eq:answer_init} to $0$ which effectively removes the query representation from the predicted answer representation. We found that the prediction changed to ``people" which can be explained by the support weights. This finding confirms our premise that the query is able to put a bias on the final answer, and that the use of the query itself maybe be beneficial for answer prediction.

\section{Conclusion}

We have presented a new type of neural network architecture for answering queries. It is end-to-end trainable and learns to utilize knowledge in the form of supporting query-answer pairs to infer the answer to a given query. It explicitly separates the query representation used for selecting support from the answer representation used for prediction. Results on recently proposed benchmark datasets for the task of reading comprehension show that our model outperforms or approaches various state-of-the-art baselines. This shows that the idea of explicitly separating query and answer is important for tasks that involve answering queries.

Future work involves the extension of this architecture to be able to properly handle other kinds of queries, e.g., list-queries or queries expecting generated answers. Furthermore, we believe that our architecture is suited for the successful application to a variety of tasks in the area of information extraction.

\section*{Acknowledgments}
We thank Thomas Demeester, Thomas Werkmeister, Sebastian Krause, Tim Rockt\"aschel and Sebastian Riedel for their comments on an early draft of this work. This research was supported by the German Federal Ministry of Education and Research (BMBF) through the projects ALL SIDES (01IW14002), BBDC (01IS14013E), and Software Campus (01IS12050, sub-project GeNIE).

\bibliography{tacl}
\bibliographystyle{acl2012}

\end{document}